\documentclass[12pt]{article}

\usepackage[margin=1in]{geometry} 

\usepackage{amsmath}
\usepackage{amsthm}
\usepackage{amssymb}

\usepackage[utf8]{inputenc}
\usepackage{url}

\usepackage[authoryear,round]{natbib}
\theoremstyle{plain}

\theoremstyle{definition}

\usepackage[dvipsnames]{xcolor}

\definecolor{safeblue}{RGB}{0,90,181}

\usepackage{graphicx, color, svg}
\usepackage{subcaption}
\graphicspath{{img/}}
\usepackage{setspace}
\usepackage{booktabs}

\usepackage{algorithm, algpseudocode} 
\usepackage{mathrsfs} 

\allowdisplaybreaks

\title{Gamifying the Vehicle Routing Problem \\ with Stochastic Requests}
\author{Nicholas D. Kullman$^1$ \and Nikita Dudorov$^2$ \and Martin Cousineau$^3$ \and Jorge E. Mendoza$^3$ \and Justin C. Goodson$^4$}

\date{
\begin{footnotesize}
	$^1$LIFAT EA 6300, CNRS, ROOT ERL CNRS 7002, Universit\'e de Tours\\%
        $^2$\'Ecole Polytechnique\\%
	$^3$Department of Operations Management and Logistics, HEC Montr\'eal\\%
        $^4$Department of Operations and Information Technology Management, Richard A. Chaifetz School of Business, Saint Louis University\\[2ex]%
\end{footnotesize}
July 4, 2024
}

\begin{document}
\onehalfspacing

\maketitle
	
\begin{abstract}
\noindent Do you remember your first video game console? We remember ours. Decades ago, they provided hours of entertainment. Now, we have repurposed them to solve dynamic and stochastic optimization problems. With deep reinforcement learning methods posting superhuman performance on a wide range of Atari games, we consider the task of representing a classic logistics problem as a game. Then, we train agents to play it. We consider several game designs for the vehicle routing problem with stochastic requests. We show how various design features impact agents’ performance, including perspective, field of view, and minimaps. With the right game design, general purpose Atari agents outperform optimization-based benchmarks, especially as problem size grows. Our work points to the representation of dynamic and stochastic optimization problems via games as a promising research direction. 
\end{abstract}

\section{Prologue}

Several of the authors are old enough to remember their very own Atari 2600. A joystick with an orange button, a black game console with a few manual switches, and a cable that connected to a 1980s television. This magical device transported us to pixelated worlds where we spent many blissful hours playing classics like Pong, Centipede, and Frogger. Thirty-five years later, during a time when gamers connect their Xbox consoles to LCDs, our childhood memories are finding new life. With the original Atari games serving as benchmarks for deep reinforcement learning (DRL) methods, we set out to bring our past into our present. As adults, we spend our professional lives wading through our own pixelated world of integer variables, discrete state spaces, and discontinuous functions. Instead of proving theorems and devising algorithms to solve dynamic and stochastic optimization problems, what if we played them like a video game? Can we map one pixelated world to another and leverage DRL to crack the code? This paper is about our quest to defeat one such problem, with only a vintage joystick and modern AI.


DRL methods have been applied to a variety of tasks involving sequential decisions and uncertainty. These tasks span the domains of healthcare \citep{Liu2017}, image recognition \citep{choi2018real}, and autonomous driving \citep{Sallab2017}, to name a few. In the operational realm, members of our team have applied DRL to dynamic taxi dispatching \citep{Kullman2019}, Amazon has used DRL for online bin packing, newsvendor, and vehicle routing problems \citep{Balaji2019}, and others have applied it to production scheduling \citep{Palombarini22, Xinquan23}. Perhaps the most well-known application is to games, where DRL has achieved superhuman performance in chess \citep{Silver2018}, Go \citep{Silver2018}, Doom \citep{Lample2017}, Texas Hold'em Poker \citep{Heinrich2016}, and StarCraft II \citep{Vinyals2019}. Notably, the architecture of \citet{Mnih2015} outperforms humans on the majority of 49 Atari games. This is achieved despite differences across the suite of games, such as appearance, goals, rewards, and actions. Indeed, it was the success of \citet{Mnih2015} that led to our reflection on whether similar DRL methods might perform comparably on any game with a related format. Games, like dynamic and stochastic optimization problems, are challenging because they involve long sequences of decisions in the face of uncertain outcomes. Thus, one also wonders if game-based representations of such problems, paired with DRL, can lead to viable solution methodologies. The possibility of bridging these two worlds motivates this paper.

To test our hypothesis, we consider how to gamify the \emph{vehicle routing problem with stochastic requests} (VRPSR). The VRPSR is an important problem in modern logistics. It is the problem of dynamically routing a vehicle to service customer requests that occur at random times across an operating horizon and in random places within a service area. The objective is to identify a routing policy that maximizes the expected number of serviced requests. The VRPSR models a range of operational scenarios, including those faced by service technicians, meal delivery services, and couriers. Identifying an optimal VRPSR policy is challenging, especially for instances of any practical size. If the VRPSR can be represented as a game, and if DRL can provide a useful solution methodology, then this would be a notable achievement. Rather than relying on tailored optimization routines to aid online decision making, academics and practitioners alike could instead turn to a more general DRL method.

\begin{figure}
    \centering
    \includegraphics[scale=0.55]{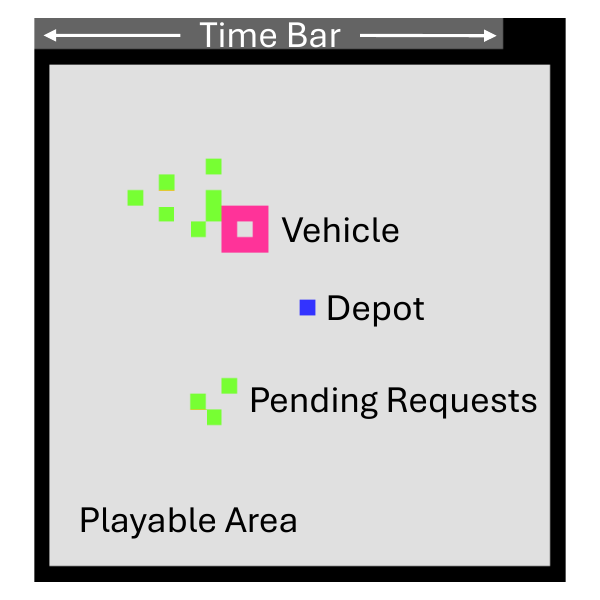}
    \caption{The VRPSR Game World}
    \label{fig:gameworld}
\end{figure}

We explore the benefits and drawbacks of various designs for the VRPSR game. Each design presents the player with a different view of the game world shown in Figure~\ref{fig:gameworld}, and thus provides different input to the DRL algorithm. First, we consider perspective, whether to reposition the vehicle in response to joystick-style movements, or to fix the vehicle in the center of the view and reposition the playable area. Second, we investigate wide and narrow fields of vision, which determine whether the entire playable area is visible to the player at once, or only some portion of it. Third, we examine the impact of including a minimap in the view. When the player's field of vision is narrow, this feature provides the player with a downscaled overview of the entire playable area.

We train agents on each view via the DRL method of \citet{Bellemare2017}, which builds on the success of \citet{Mnih2015}, then execute them across various VRPSR problem instances. Computational experiments yield four conclusions. First, fixing the vehicle in the center of the view leads to better performance than moving the vehicle within a static view. The improvement follows from an agent's ability to associate each pixel in the view with movement in a particular direction, rather than with a function of the vehicle's location in a static view. Second, a narrow field of vision is helpful for larger playable areas. This feature shifts the problem of learning across a larger playable area from an issue of more complex input to one of additional exploration. Third, including a minimap in the view further improves agent performance. The two features in tandem combine local resolution across a narrow field of vision with a global approximation of the entire playable area. Fourth, in comparison to optimization-based benchmarks, DRL agents perform better as the expected number of requests increases. Because the benchmarks make computations on the fly, the requirement to make timely decisions limits their ability to select good actions. In contrast, following offline training, DRL agents require little time to make online decisions. 

Though some of our game designs allow agents to outperform benchmark policies, our aim is not to develop a state-of-the-art procedure. Rather, our contribution is a connection between the seemingly disparate worlds of video games and logistics. More generally, our work points to the representation of dynamic and stochastic optimization problems via games as a promising research direction.

The paper proceeds as follows. In \S\ref{sec:earlier}, we review related literature. In \S\ref{sec:gameworld}, we detail the game setup and describe four game views. In \S\ref{sec:hero}, we describe the DRL training procedure. In \S\ref{sec:god}, we present benchmark methods. In \S\ref{sec:play}, we conduct computational experiments and discuss their implications. We conclude the paper in \S\ref{sec:conc}.

\section{Earlier Versions}\label{sec:earlier}


It seems fitting that the start of VRPSR research coincides with the release of the Atari 2600. The work of \citet{psaraftis1980dynamic} reoptimizes a route through pending requests whenever a new request is made, then uses the route to direct vehicle movement. Just like the Atari influenced video game design for years to come, \citeauthor{psaraftis1980dynamic} inspired similar research across several decades. \citet{gendreau1999parallel}, \citet{ichoua2000diversion}, \citet{van2004dynamic}, \citet{gendreau2006neighborhood}, \citet{Ichoua06}, \citet{Branchini09}, and \citet{ferrucci2013pro} all present variations on the original idea of Psaraftis. Innovations center on more advanced routing heuristics. While these methods exploit advances in deterministic routing, for the most part they do not explicitly consider future requests.

More anticipatory VRPSR decision making begins with \citet{Bent04SBP}. Rather than direct the vehicle based on reoptimization of a single route, \citeauthor{Bent04SBP} reoptimize a collection of routes, each of which contains a different random sample of future requests. A consensus function then determines vehicle movement. \citet{Hvattum06} and \citet{ghiani2009anticipatory} proceed similarly. Concurrently, \citet{Mitro04}, \citet{branke2005waiting}, \citet{thomas2007dynamic}, and \citet{ghiani2012comparison} explore waiting strategies. These methods dynamically move and halt the vehicle in anticipation of future requests.

The work of \citet{meisel2011anticipatory} begins an era of VRPSR value function approximation, which gives explicit consideration to the timing and locations of future requests. Building on \citet{meisel2011anticipatory}, \citet{ulmer2018a} aggregates states around time-based features, then dynamically partitions the space as part of an approximate value iteration procedure. \citet{ulmer2018b} extends these ideas to a multi-period VRPSR and \citet{ulmer2018offline} combines them with rollout algorithms. Even though the method of \citet{ulmer2018a} adapts to the approximation process, it cannot revert partitioning decisions once made. The adaptive procedure of \citet{soeffker19} remedies this issue and yields even better value function approximations.

More recently, VRPSR value function approximations are made via neural networks. \citet{joe20} show that neural nets can outperform the scenario-based method of \citet{Bent04SBP} as well as an approximate value iteration procedure. In the context of same-day delivery, \citet{chen21} use neural networks to learn the value of fulfilling a request via drone versus via truck. Our approach also relies on neural network approximations of the value function. However, in contrast to the literature, we do not design a value function approximation for the VRPSR. Rather, we seek to design a game-based representation of the VRPSR for a preexisting network architecture. For a more comprehensive examination of dynamic and stochastic vehicle routing literature, see \citet{soeffker2022}.

\section{The Game World}\label{sec:gameworld}

The vehicle routing problem with stochastic requests (VRPSR) dispatches a single vehicle to meet customer requests arriving at random times across a given operating horizon and at random locations across a known service area. The objective is to design a dynamic routing policy, beginning and ending at a depot, that maximizes the expected number of serviced customers. The \emph{game world}, portrayed in Figure~\ref{fig:gameworld}, is a visual representation of the problem description. It is composed of several elements. The \emph{playable area} represents the service area, or the portion of the game world within which the vehicle may move. Within the playable area, the depot and requests are represented by single pixels. Customers that have requested service are green and the depot is blue. The pixel depicting each request is invisible before the time of request and after service. The vehicle’s location is shown by the open pixel in the center of the pink square. The vehicle services a request by navigating its open pixel to a customer's position. The playable area is surrounded by a thin border. Above the border is a rectangular \emph{time bar} whose length represents the remaining time before the vehicle must return to the depot. 

The VRPSR is conventionally formulated as a Markov decision process: epochs occur when customers are serviced and when customers request service; the state tracks time, vehicle position, and the locations of pending requests; actions direct the vehicle to a pending service request or to wait at the current location; rewards count the number of serviced customers; and transitions account for the likelihood of new requests across space and time. Formal models for related problems may be found in \citet{ulmer2018offline} and \citet{routemdp2020}.

\begin{figure}
    \centering
    \includegraphics[width=6.3in]{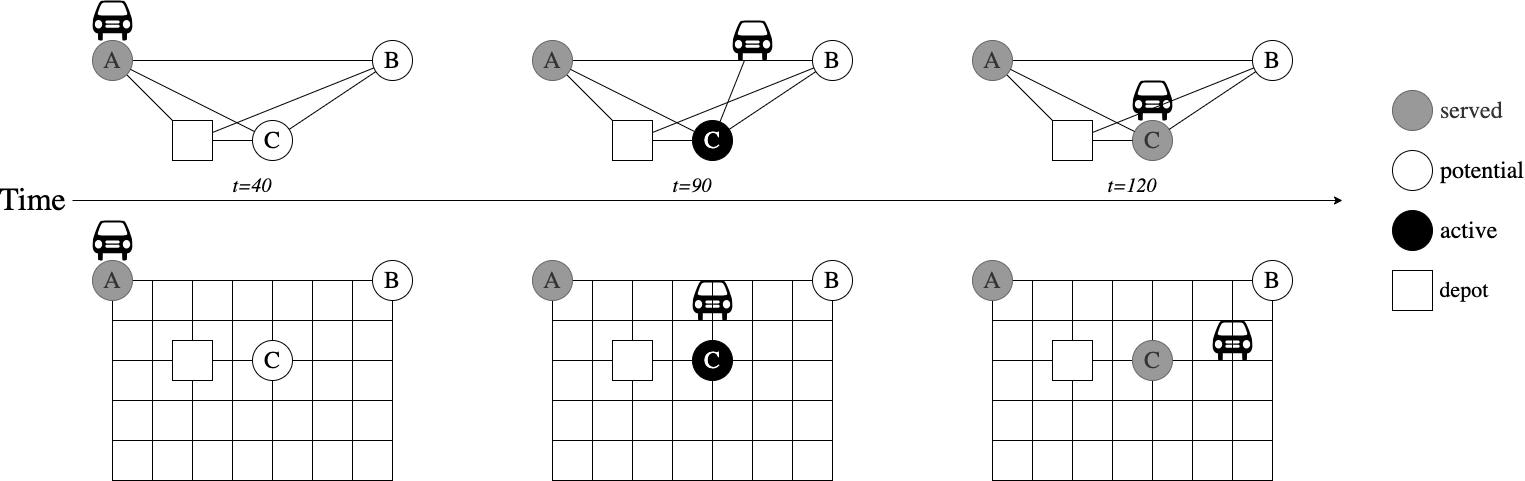}
    \caption{Graphs for the traditional (top) and game (bottom) representations of the VRPSR.}
    \label{fig:graphs}
\end{figure}

Modeling the VRPSR as a game requires modifications to how policies route the vehicle and to the conventional formulation. The operations research community typically considers policies that move the vehicle among the edges of a complete graph $G$ consisting of nodes for each customer and the depot. To accommodate the pixel-by-pixel movements typical to games, routing policies in the VRPSR game move the vehicle along the edges of a graph $G^\prime$ representing the playable area. Each node in $G^\prime$ is a pixel and edges connect a pixel to adjacent pixels, e.g., the pixels above, below, and to either side. While routing across $G$ allows for a variety of distance metrics (e.g., Euclidian), routing across $G^\prime$ requires Manhattan-style movements. Figure~\ref{fig:graphs} depicts the VRPSR on both graphs.

A game formulation of the VRPSR modifies actions, epochs, and states. The action space consists of at most five actions: move up, move down, move left, move right, and no movement. If the vehicle is at the boundary of the playable area, the number of feasible actions is fewer. In the conventional formulation, the path between any pair of nodes in $G$ incurs the minimum travel time. In contrast, moving from one customer to another in $G^\prime$ can be accomplished via many paths. For preemptive routing policies, movement across $G^\prime$ offers more flexibility than movement across $G$. For example, though detouring the vehicle away from a shortest path may result in more travel time, it may also put the vehicle in closer proximity to potential requests, thus making it possible to service additional requests while en route to a customer. Figure~\ref{fig:graphs} depicts this behavior. In the top part of the figure, the vehicle moves directly from node A to node B through graph $G$. In the bottom portion, movement toward node B through graph $G^\prime$ anticipates the possibility of a request at node C by moving toward B along a lower and longer path. Epochs reflect the additional routing flexibility afforded by movement across $G^\prime$. They occur at each moment of a discretized timeline.

\begin{figure}
     \centering
     \begin{subfigure}[b]{0.29\textwidth}
         \includegraphics[width=\textwidth]{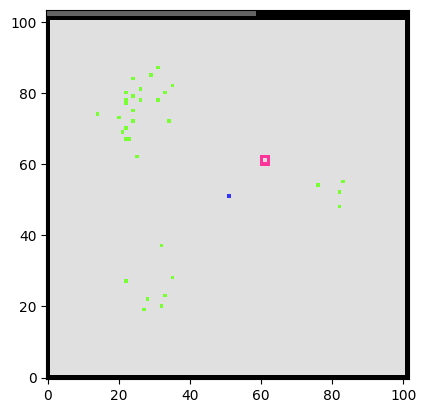}
         \caption{}
         \label{fig:view_standard}
     \end{subfigure}
     \begin{subfigure}[b]{0.29\textwidth}
         \includegraphics[width=\textwidth]{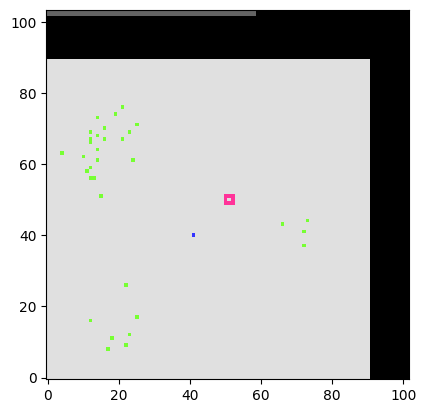}
         \caption{}
         \label{fig:view_centered_car}
     \end{subfigure}
     \vfill
     \begin{subfigure}[b]{0.29\textwidth}
         \includegraphics[width=\textwidth]{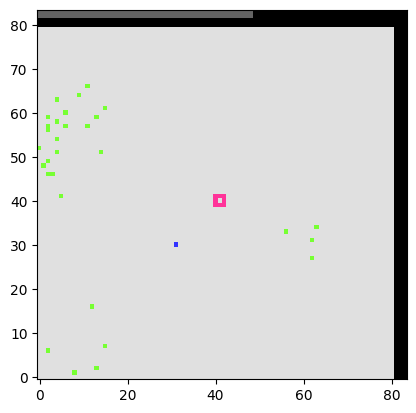}
         \caption{}
         \label{fig:view_zoom}
     \end{subfigure}
     \begin{subfigure}[b]{0.29\textwidth}
         \includegraphics[width=\textwidth]{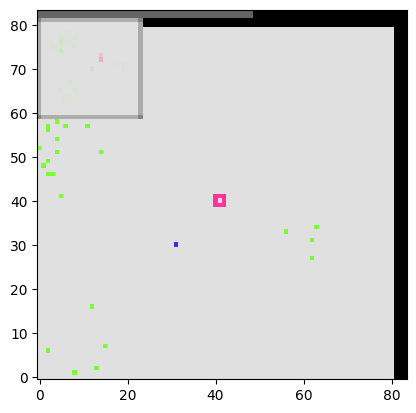}
         \caption{}
         \label{fig:view_zoom_map}
     \end{subfigure}
     \caption{Game views for a 100-by-100 pixel playable area: (a) World View (b) Vehicle View (c) Zoom View (d) Survey View. Zoom and Survey Views are depicted with an 84-by-84 pixel field of vision.}
     \label{fig:view}
\end{figure}

The state of a VRPSR game is the player's \emph{view} of the game world at an epoch. A view consists of the time bar plus the player's \emph{field of vision}. The player's field of vision is the visible portion of the playable area. We consider four views, each of which is depicted in Figure~\ref{fig:view}. In the \emph{World View}, the field of vision includes the entire playable area and the player simply moves the vehicle within the area. In the \emph{Vehicle View}, the vehicle is fixed in the center of the field of vision and movement adjusts the position of the field of vision across the playable area. When the field of vision is positioned away from the center of the playable area, portions of the playable area move out of the view. The \emph{Zoom View} is the Vehicle View with a narrower field of vision, meaning portions of the playable area are obscured even when the vehicle is positioned in the center of the playable area. To observe portions of the playable area outside the field of vision, the vehicle must move to those regions. The \emph{Survey View} is the Zoom View plus a minimap. In this view, players see a small overview of the playable area in the top left corner. The overview is an aggregation, where the color of each aggregated pixel is the average color of the pixels it represents. 

Because each view is a different state representation, the four views constitute four different VRPSR games. While the World View provides a complete observation of the state variable, the latter three views result in partial observations.


\section{Our Hero}\label{sec:hero}

We use the \emph{categorical Deep $Q$-Network} (DQN) approach of \citet{Bellemare2017} to train agents for all four views of the VRPSR game. This method has demonstrated particular success across a suite of Atari 2600 games. In contrast to classical reinforcement learning techniques, which estimate expected $Q$-values, categorical DQN approximates $Q$-value distributions. The method builds on the celebrated DQN architecture of \citet{Mnih2015}, which employs a collection of neural network techniques. As in \citet{Mnih2015}, categorical DQN takes the classical $\epsilon$-greedy approach to learning. However, instead of minimizing squared loss between $Q$-values, it minimizes sample loss between distributions of $Q$-values. 

\begin{figure}
    \centering
    \includegraphics[scale=0.2]{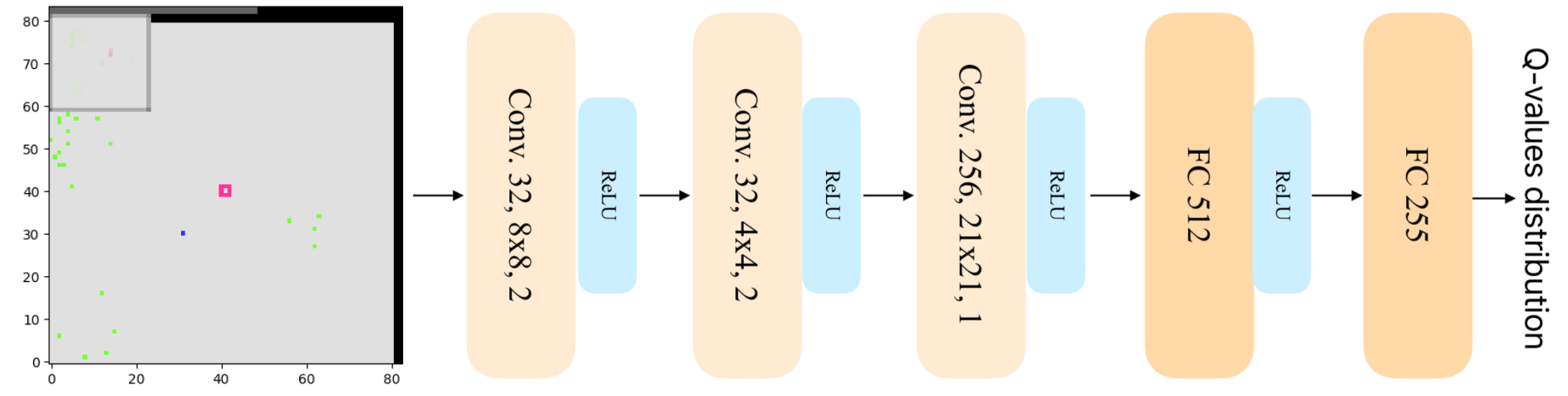}
    \caption{Categorical DQN Architecture}
    \label{fig:dqn_architecture}
\end{figure}

Our network architecture is shown in Figure~\ref{fig:dqn_architecture}. It takes as input a game view, represented as an array of pixels. It consists of three convolution layers and two fully connected layers. The architecture mirrors that of \citet{Bellemare2017}, with one modification. Because each customer occupies only a single pixel, a customer's exact location may become lost in the convolution layers. To prevent this, in the first convolution layer we double the number of filters and cut the stride in half. After an agent is trained, $Q$-value distributions are used to make decisions by selecting an action that maximizes the expected $Q$-value in a given state.

\section{God Mode and Other Players}\label{sec:god}

We benchmark agents' performance against two reoptimization policies and the expected value with perfect information (EVPI), which serves as a dual bound on the value of an optimal policy \citep{Brown10}. At a given state, both policies select an action by solving a mixed-integer linear program (MILP). The MILP seeks a route that serves a maximum number of pending service requests. The action is movement of the vehicle toward the first customer visited by the route. We refer to these policies as reoptimization with preemption (ReoptWP) and reoptimization without preemption (ReoptWOP). The ReoptWP policy solves the MILP when customers are serviced and whenever service requests are realized. Thus, a vehicle en route to one customer can be diverted to a different customer. The ReoptWOP policy only solves the MILP when customers are serviced, as is customary in the VRPSR literature. This policy commits the vehicle to the selected request. ReoptWP and ReoptWOP may be viewed as players of the VRPSR game with a World View. As discussed in \S\ref{sec:earlier}, solving deterministic MILPs in rolling horizon fashion to make dynamic decisions is a common means of deriving policies from static methods. The policies provide a sense of how classical optimization-based methods perform relative to a state-of-the-art reinforcement learning method. As we discuss below, the same MILP can be used to obtain the EVPI.

To formalize the MILP, let $\bar{t}$ be the current time and let node $n$ in $G^\prime$ be the location of the vehicle at time $\bar{t}$. Denote by $C \subset G^\prime$ the subset of nodes in $G^\prime$ corresponding to customers who have requested service but who have not yet been visited. Let $0$ in $G^\prime$ represent the depot and denote by $\bar{C} = C \cup \{n\} \cup \{0\}$ the set of nodes comprising customer locations, vehicle location, and the depot. For each pair of nodes $(i,j)$ in $\bar{C} \times \bar{C}$, let $d_{ij}$ be the duration of the shortest travel time from $i$ to $j$, where distances are Manhattan. Denote by $r_i$ the time of the request at node $i$ in $C$. Let $l_i = T - d_{i0}$ be the latest time the request at node $i$ in $C$ may be serviced such that the vehicle can return to the depot by the end of the operating horizon at time $T$.

Decision variables include which customers to visit and when. Let $h_i$ be $1$ if the request at node $i$ is serviced by the route and $0$ otherwise. Let $y_i$ be $1$ if the request at node $i$ is the first request to be serviced and $0$ otherwise. Let $x_{ij}$ be $1$ if the request at node $j$ is serviced immediately after the request at node $i$ and $0$ otherwise. Finally, let $t_i$ be the time at which the request at node $i$ is serviced, if it is serviced at all. Constraints require the vehicle to begin routing from its current location, to visit customers at or after the time at which service is requested, to visit each customer at most once, and to conclude service with enough time to return to the depot before the end of the operating horizon.

The MILP is formulated as follows:
\begin{align}
\text{maximize}\quad & \sum_{i \in C} h_i \label{eqn:1} \\
\text{subject to}\quad & \sum_{i \in C} y_i \leq 1 \label{eqn:2} \\
& y_j + \sum_{i \in C} x_{ij} = h_j, \quad j \in C \label{eqn:3} \\
& \sum_{j \in C} x_{ij} \leq h_i, \quad i \in C \label{eqn:4} \\
& t_i \geq r_i + \left( \bar{t} + d_{ni} - r_i \right) y_i, \quad i \in C \label{eqn:5} \\
& t_j - t_i \geq r_j - l_i + \left( d_{ij} - r_j + l_i \right) x_{ij}, \quad (i,j) \in C \times C \label{eqn:6} \\
& r_i \leq t_i \leq l_i, \quad i \in C \label{eqn:10} \\
& h_i \in \{0,1\}, \quad i \in C \label{eqn:7} \\
& y_i \in \{0,1\}, \quad i \in C \label{eqn:8} \\
& x_{ij} \in \{0,1\}, \quad (i,j) \in C \times C \label{eqn:9}
\end{align}
The objective in Equation~\eqref{eqn:1} seeks to maximize the number of serviced requests. Constraint~\eqref{eqn:2} allows at most one request to be routed first. Constraints~\eqref{eqn:3} allow each node $j$ in $C$ to be routed only if $j$ is selected for service. Constraints~\eqref{eqn:4} allow departure from node $i$ in $C$ only if node $i$ is marked for service. If the request at node $i$ in $C$ is routed first, then Constraints~\eqref{eqn:5} require the time of service at node $i$ to be at or after the current time plus the travel time from the vehicle's current location to node $i$. If the request at node $j$ in $C$ is routed after the request at node $i$ in $C$, then Constraints~\eqref{eqn:6} require the time of service at node $j$ to be at or after the time of service at node $i$ plus the travel time from node $i$ to node $j$. Constraints~\eqref{eqn:10} ensure that service at each node $i$ in $C$ occurs between $r_i$ and $l_i$. Constraints~\eqref{eqn:7} and \eqref{eqn:8} mark decision variables $h_i$ and $y_i$ as binary for each $i$ in $C$. Similarly, Constraints~\eqref{eqn:9} require $x_{ij}$ to be binary for each pair of nodes in $C \times C$.

The ReoptWP and ReoptWOP policies extract actions from feasible MILP solutions as follows. Let $i^\star$ be the node in $C$ such that $y_{i^\star} = 1$. Node $i^\star$ is the first request serviced by the solution. Both ReoptWP and ReoptWOP advance the vehicle toward $i^\star$ on a path that first moves horizontally and then moves vertically. If $i^\star$ does not exist and $\bar{t} + d_{n 0} < T$, then the action is to wait at $n$, the vehicle's current location. Otherwise, the action moves the vehicle along a shortest path to the depot.

The EVPI is the expected value of an optimal policy with perfect foresight of the times and locations of service requests. If we represent a realization of requests by $C$ and their times by $r = (r_i)_{i \in C}$, then the MILP models the problem of identifying an optimal routing in response to $C$ and $r$. Denote by $f(C,r)$ the value of this routing. The EVPI is the expected value $\mathbb{E}[f(C,r)]$ across all possible realizations.

\section{Let's Play}\label{sec:play}

In this section, we conduct three sets of experiments designed to demonstrate the strengths and weaknesses of agents trained on each of the four game views. Problem instances and algorithmic parameters are described in \S~\ref{sec:loading}. Results are presented in \S~\ref{sec:results}.

\subsection{Loading} \label{sec:loading}

Across all experiments, instances are structured similar to those of \citet{ulmer2018offline}. Instances for the first set of experiments are characterized by a 20 km square service region, a 6-hour operating horizon, and 30 expected requests. A 100-by-100 pixel playable area represents the service region, where each pixel is 0.2 km square. Time is discretized into 750 steps, each with a duration of 0.008 hours. This leads to 749 epochs plus a final time step that concludes service. The vehicle moves at a constant speed of 25 kph, or one pixel per time step. The depot is located in the center of the playable area. The number and location of requests are independent random variables. The number of customers that request service in a given time step is Poisson distributed with rate $30/749$, resulting in 30 expected requests across the operating horizon. Each request is located in one of three areas with likelihoods 0.25, 0.50, and 0.25, respectively. Request locations within each area follow a bivariate normal distribution. Referencing the bottom-left corner of the playable area as the origin, mean locations for the areas are $(5 \text{ km}, 5 \text{ km})$, $(5 \text{ km}, 15 \text{ km})$, and $(15 \text{ km}, 10 \text{ km})$, respectively. Standard deviations for both horizontal and vertical dimensions are 1 km and the correlation between dimensions is 0.

The second set of experiments modifies instances for the first set by increasing the resolution of the playable area from 100-by-100 pixels to 200-by-200 pixels. With the area of each pixel at 0.1 km square, time is discretized into 1500 steps, each with a duration of 0.004 hours. The Poisson rate parameter is adjusted accordingly to $30/1499$. The third set of experiments alters instances from the first set by increasing the expected number of customers from 30 to 100. All other parameters remain the same.


We use categorical DQN to train agents on VRPSR games via the World, Vehicle, Zoom, and Survey Views. We refer to the agents by the views on which they are trained. The field of vision for the Zoom and Survey Views is 84-by-84 pixels, the same resolution as the images used by \citet{Mnih2015}. To accelerate exploration, each training episode begins by moving the vehicle to a random location. In the first and third sets of experiments, each agent is trained for 50 million epochs, with episodes terminating whenever the remaining time is such that the agent’s only feasible moves are those bringing it back to the depot. In the second set of experiments, the number of training epochs increases to 180 million. We use \citet{rllib} to implement categorical DQN. It provides the pre-tuned hyperparameters for Atari displayed in Table~\ref{tab:params}. Training is conducted in serial on Nvidia V100 GPUs.


\begin{table}
  \centering
  \caption{Ray RLlib Hyperparameter Values}
  \label{tab:params}
    \begin{tabular}{rl}
    \toprule
    \textbf{Parameter} & \textbf{Value} \\
    \midrule
    num\_atoms & 51 \\
    replay\_buffer\_config: capacity & 1,000,000 \\
    target\_network\_update\_freq & 8,000 \\
    lr    & 0.0000625 \\
    adam\_epsilon & 0.00015 \\
    hiddens & [512] \\
    train\_batch\_size & 32 \\
    exploration\_config: epsilon\_timesteps & 200,000 \\
    exploration\_config: final\_epsilon  & 0.01 \\
    \bottomrule
    \end{tabular}%
  \label{tab:addlabel}%
\end{table}%

The expected values of agents' decisions and of each benchmark policy are estimated via simulation. We do this by randomly generating request realizations across the operating horizon and service area, executing the policy, recording the number of serviced requests, then repeating a total of 250 times. The average number of serviced requests is an unbiased and consistent estimator of the expected number of serviced requests. For each set of request realizations, the ReoptWOP policy is allocated a computing budget equal to the 6-hour operating horizon. In the first and second sets of experiments, this provides up to 12 minutes to solve the MILP for each of 30 expected requests. In the third set of experiments, ReoptWOP receives up to 3.6 minutes to solve the MILP for each of 100 expected requests. The ReoptWP policy receives the same allocations for each MILP. If a MILP is not solved to optimality within the allocated time, then the best feasible solution is used to direct action selection. The EVPI is estimated via simulation across the same 250 instances. For each realization of requests, each MILP instance is allocated 6 hours of computing time. If the solver does not identify an optimal solution to the MILP within that time, then we use the solver's smallest upper bound instead. Though this may overestimate the EVPI, it is also a dual bound on the value of an optimal policy. All MILPs are solved with Gurobi Optimizer on Intel Xeon Gold 6148 (2.4 GHZ) CPUs. The experimental setup is summarized in Table~\ref{tab:exp}.

\begin{table}[htbp]
  \centering
  \caption{Experimental Setup}
    \begin{tabular}{l|ccc}
    \toprule
    \textbf{Experiment Set} & \textbf{1}     & \textbf{2}     & \textbf{3} \\
    Expected Requests & 30    & 30    & 100 \\
    Playable Area (pixels square) & 100   & 200   & 100 \\
    World and Vehicle Views Field of Vision (pixels square) & 100    & 200    & 100 \\
    Zoom and Survey Views Field of Vision (pixels square) & 84    & 84    & 84 \\
    Training Iterations per Agent (millions) & 50    & 180   & 50 \\
    ReoptWOP and ReoptWP CPU Minutes per MILP & 12    & 12    & 3.6 \\
    \bottomrule
    \end{tabular}%
  \label{tab:exp}%
\end{table}%


\subsection{Gameplay} \label{sec:results}

\begin{figure}
    \centering
    \includegraphics[scale=0.8]{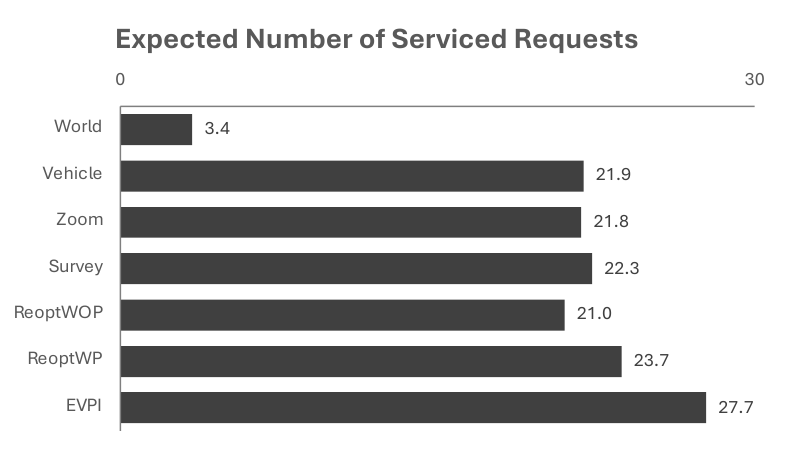}
    \caption{Performance in the 100-by-100 playable area with 30 expected requests}
    \label{fig:resultsall}
\end{figure}

\begin{figure}
    \centering
    \includegraphics[width=0.7\textwidth]{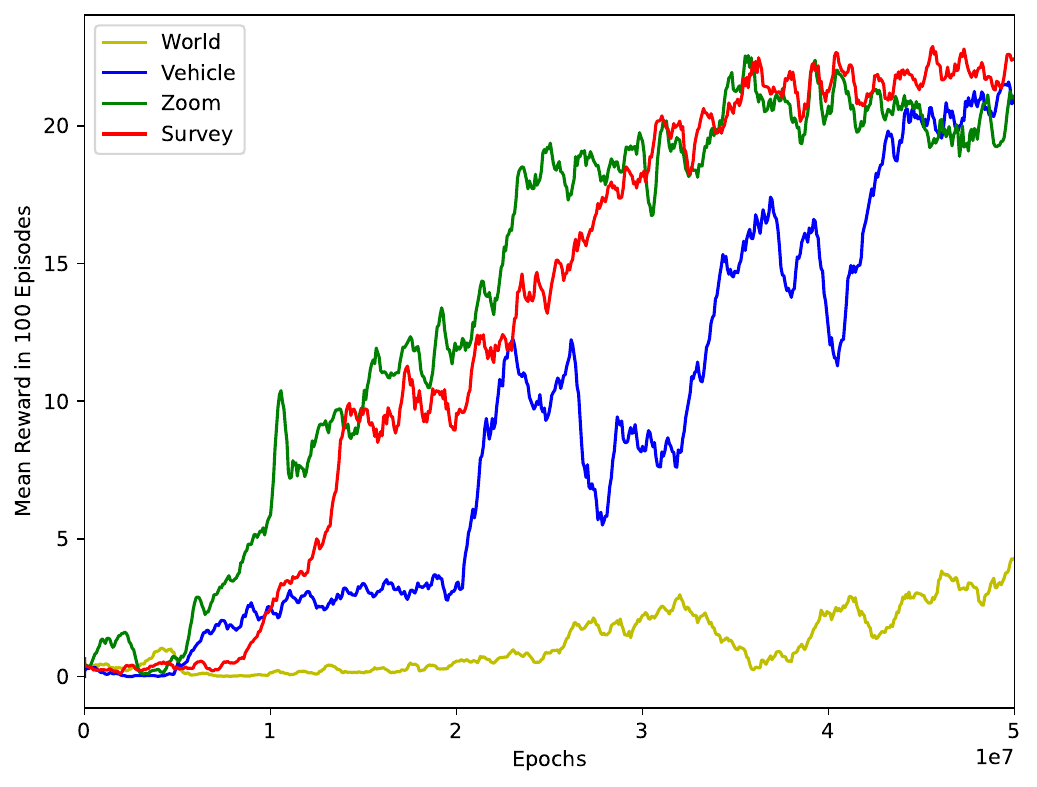}
    \caption{Training curves for the 100-by-100 playable area with 30 expected requests}
    \label{fig:views_comparison}
\end{figure}

The first set of experiments highlights the benefit of a Vehicle View. Figures~\ref{fig:resultsall} and~\ref{fig:views_comparison} depict the results. Figure~\ref{fig:resultsall} displays the performance of each agent and of the benchmarks while Figure~\ref{fig:views_comparison} shows the performance of each agent as a function of the number of training epochs. The expected number of serviced requests by the World Agent (3.4) is particularly low. The Vehicle (21.9), Zoom (21.8), and Survey Agents (22.3) each demonstrate substantial improvements over the World Agent and also outperform the ReoptWOP policy (21.0). While these three agents perform similarly, the Survey Agent's zoom and minimap features together provide an edge over the Vehicle View in isolation. Figure~\ref{fig:views_comparison} underscores these results. While all four agents improve as the number of training epochs approaches 50 million, the World Agent lags behind the other three, and the two agents trained on the zoom feature surpass the performance of the Vehicle Agent. Although the ReoptWP policy (23.7) posts the best performance, the Survey Agent is not far behind. The EVPI (27.7) sits between the expected number of requests serviced by ReoptWP and 30, the expected number of total requests. Of the 250 instances used to calculate the EVPI, nearly 25 percent solve to optimality within the allotted time. Across the vast majority of the remaining instances, the solver reports optimality gaps below 20 percent.


The performance disparity between the World Agent and the Vehicle Agent points to a Vehicle View as an important factor in modeling the VRPSR as a game. Vehicle Views allow agents to associate each pixel in the view with specific actions, e.g., pixels in the top half of the view require upward movement and pixels on the right side of the view require movement to the right. In contrast, the action required to move the vehicle to the same pixel in the World View is a function of the vehicle's location in the view. The difference is one of gauging value relative to the view versus value relative to the vehicle's position in the view. It seems that dependency on vehicle location introduces non-trivial complexity. While it is straightforward for humans to recognize the connection between the World View and the Vehicle Views, to the deep $Q$-networks that drive the categorical DQN method, Vehicle Views are more amenable to learning.

\begin{figure}
    \centering
    \includegraphics[scale=0.8]{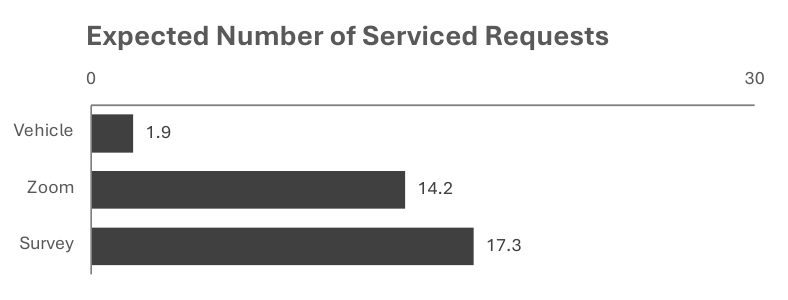}
    \caption{Performance in the 200-by-200 playable area with 30 expected requests}
    \label{fig:results200}
\end{figure}

\begin{figure}
    \centering
    \includegraphics[width=0.7\textwidth]{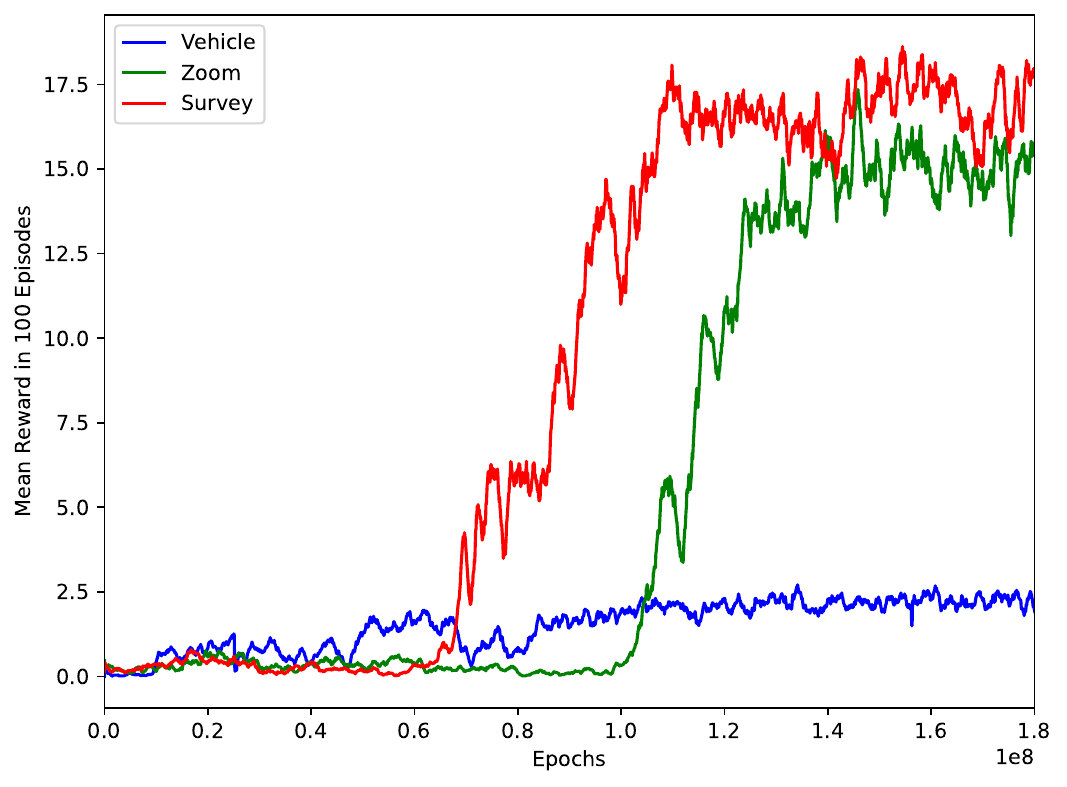}
    \caption{Training curves for the 200-by-200 playable area with 30 expected requests}
    \label{fig:views_comparison1}
\end{figure}

The second set of experiments showcases the benefit of Zoom Views and confirms the utility of the Survey View. Figures~\ref{fig:results200} and~\ref{fig:views_comparison1} depict the results. The format of these figures mirrors that of Figures~\ref{fig:resultsall} and~\ref{fig:views_comparison}. The World Agent is excluded from these experiments. Because the increased resolution does not affect the benchmarks, their performance values are not displayed. Relative to the first set of experiments, the expected number of serviced requests decreases for each agent. However, the Zoom (14.2) and Survey Agents (17.3) both improve substantially over the Vehicle Agent (1.9). The training curves in Figure~\ref{fig:views_comparison1} emphasize this difference. Across 180 million training epochs, the Vehicle Agent demonstrates only minor performance gains whereas the other two agents exhibit marked improvement.

Theses results indicate that as view resolution increases, performance of the agents decreases. For the Vehicle Agent, quadrupling the number of pixels from 100-by-100 to 200-by-200 quadruples the size of the input array for the deep $Q$-networks that underlie the categorical DQN method. As demonstrated by the training curves in Figure~\ref{fig:views_comparison1}, this is a more difficult learning task. The zoom feature included in the Zoom and Survey Agents partially offsets the challenges of a larger playable area by maintaining the size of the view at 84-by-84 pixels. This shifts the problem of learning across a larger playable area from an issue of more complex input to one of additional spatial exploration. In the first set of experiments, this difference is not evident because the Zoom View captures more than 70 percent of the playable area. In the second set of experiments, the Zoom View captures four times less, not even 18 percent, and leads to substantially better performance. 

In both the first and second sets of experiments, the performance of the Survey Agent exceeds that of the other agents. This points to the minimap feature as an important part of VRPSR game design. In contrast to the zoom feature, which provides high resolution locally, the minimap feature offers a global approximation of the entire playable area. Combining these features into one view provides the Survey Agent with detail to facilitate exploitation and with a wide field of vision to promote exploration.

The contrast to conventional optimization-based approaches is also notable. While the performance of our agents depends heavily on the granularity of the playable area, methods like ReoptWOP and ReoptWP do not. Although agent performance is sensitive to view resolution, as we show next, it is robust to increases in problem size.



\begin{figure}
    \centering
    \includegraphics[scale=0.8]{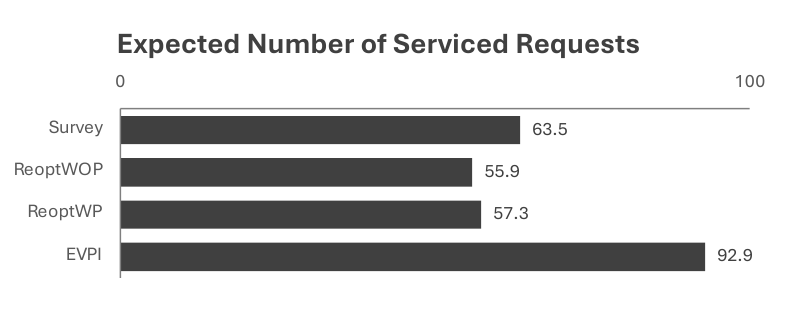}
    \caption{Performance in the 100-by-100 playable area with 100 expected requests}
    \label{fig:results100}
\end{figure}

The third set of experiments spotlights advantages of DRL when the problem size grows. These experiments examine the performance of the Survey Agent alongside the benchmarks when the expected number of requests more than triples from 30 to 100. Figure~\ref{fig:results100} depicts the results. In contrast to the first and second sets of experiments, the expected number of requests serviced by the Survey Agent (63.5) exceeds those of ReoptWOP (55.9) and ReoptWP (57.3). As a percentage of expected total number of requests, the EVPI (92.9) is comparable to its value with 30 customers.

Increasing the problem size highlights an important difference between the Survey Agent and the benchmark policies. Because a larger number of expected requests leads to appreciably larger MILPs, the ReoptWOP and ReoptWP policies struggle to select good actions within the allotted computing time of 3.6 minutes per MILP. Significant increases in computing time may lead to better policy performance, but this is not practical. The ReoptWOP and ReoptWP policies operate online. Because operational limitations typically require decisions to be made in a matter of minutes, this precludes the possibility of allocating more computing time to solving MILPs on the fly. In contrast, the Survey Agent may be trained offline for large periods of time without negative service consequences. Then, when it is deployed, actions are selected in a matter of seconds. Thus, as problem size grows, DRL yields better performance in typical operating conditions.


\section{Game Over}\label{sec:conc}

A pixelated ``Game Over'' was the message that flashed across our CRT televisions when we ran out of lives. We would start again, play until the next ``Game Over,'' and repeat until our mothers told us our eyeballs would turn into squares if we played any longer. The same ``Game Over'' also signaled the conclusion of a game. It marked completion of every level. Sometimes this unlocked new content, and sometimes we loaded a new cartridge into the console. In any case, as long as we kept playing, the game was never really over.

Game worlds, playable areas, zoom views, minimaps, deep-$Q$ networks. These things are not the end. This is only level one. New quests await us. Games, like optimization problems, have always been about discovery of a winning policy. Whether that happens through theorems and proofs, or by sitting an agent in front of a screen through millions of ``Game Overs,'' the goal is the same. This paper shows that it is possible to represent the VRPSR as a game and to train a neural network to play it better than some optimization algorithms. To keep going down this road, we must find new connections between games and optimization problems. Any number of vehicle routing problem variants could be gamified. Visual representations of scheduling, production, and assortment problems could lead to new designs. And a host of problems outside the operational domain might also be amenable to gamification. 

When we started this research, it felt a bit unusual. Our university coursework covered methods to solve equations and construct algorithms. Game design wasn’t part of the curriculum. In hindsight, however, joining our pixelated past with our pixelated present seems natural. A joystick and a game console are just new ways to think about objectives, constraints, and decision variables. Where we go from here depends on how you want to play.

\section*{Credits}
This research was enabled in part by support from Calcul Québec (\url{calculquebec.ca}), the Digital Research Alliance of Canada (\url{alliancecan.ca}), HEC Montréal, and the Institute for Data Valorization (IVADO). The authors especially thank Clément Grodecoeur for his efforts in the development and prototyping of an early VRPSR game.

\bibliography{references}

\begin{thebibliography}{39}
\providecommand{\natexlab}[1]{#1}
\providecommand{\url}[1]{\texttt{#1}}
\expandafter\ifx\csname urlstyle\endcsname\relax
  \providecommand{\doi}[1]{doi: #1}\else
  \providecommand{\doi}{doi: \begingroup \urlstyle{rm}\Url}\fi

\bibitem[Balaji et~al.(2019)Balaji, Bell-Masterson, Bilgin, Damianou, Moreno~Garcia, Jain, Luo, Maggiar, Narayanaswamy, and Ye]{Balaji2019}
B.~Balaji, J.~Bell-Masterson, E.~Bilgin, A.~Damianou, P.~Moreno~Garcia, A.~Jain, R.~Luo, A.~Maggiar, B.~Narayanaswamy, and C.~Ye.
\newblock Orl: Reinforcement learning benchmarks for online stochastic optimization problems.
\newblock \emph{arXiv preprint arXiv:1911.10641}, 2019.

\bibitem[Bellemare et~al.(2017)Bellemare, Dabney, and Munos]{Bellemare2017}
M.~Bellemare, W.~Dabney, and R.~Munos.
\newblock A distributional perspective on reinforcement learning.
\newblock \emph{Proceedings of the 34th International Conference on Machine Learning}, 70:\penalty0 449--458, 2017.

\bibitem[Bent and {Van Hentenryck}(2004)]{Bent04SBP}
R.~Bent and P.~{Van Hentenryck}.
\newblock Scenario-based planning for partially dynamic vehicle routing with stochastic customers.
\newblock \emph{Operations Research}, 52\penalty0 (6):\penalty0 977--987, 2004.

\bibitem[Branchini et~al.(2009)Branchini, Armentano, and L{\o}kketangen]{Branchini09}
R.~Branchini, A.~Armentano, and A.~L{\o}kketangen.
\newblock Adaptive granular local search heuristic for a dynamic vehicle routing problem.
\newblock \emph{Computers and Operations Research}, 36:\penalty0 2955--2968, 2009.

\bibitem[Branke et~al.(2005)Branke, Middendorf, Noeth, and Dessouky]{branke2005waiting}
J.~Branke, M.~Middendorf, G.~Noeth, and M.~Dessouky.
\newblock Waiting strategies for dynamic vehicle routing.
\newblock \emph{Transportation Science}, 39\penalty0 (3):\penalty0 298--312, 2005.

\bibitem[Brown et~al.(2010)Brown, Smith, and Sun]{Brown10}
D.~Brown, J.~Smith, and P.~Sun.
\newblock Information relaxations and duality in stochastic dynamic programs.
\newblock \emph{Operations Research}, 58\penalty0 (4):\penalty0 785--801, 2010.

\bibitem[Chen et~al.(2022)Chen, Ulmer, and Thomas]{chen21}
X.~Chen, M.~Ulmer, and B.~Thomas.
\newblock Deep q-learning for same-day delivery with vehicles and drones.
\newblock \emph{European Journal of Operational Research}, 298:\penalty0 939--952, 2022.

\bibitem[Choi et~al.(2018)Choi, Kwon, and Lee]{choi2018real}
J.~Choi, J.~Kwon, and K.~M. Lee.
\newblock Real-time visual tracking by deep reinforced decision making.
\newblock \emph{Computer Vision and Image Understanding}, 171:\penalty0 10--19, 2018.

\bibitem[Ferrucci et~al.(2013)Ferrucci, Bock, and Gendreau]{ferrucci2013pro}
F.~Ferrucci, S.~Bock, and M.~Gendreau.
\newblock A pro-active real-time control approach for dynamic vehicle routing problems dealing with the delivery of urgent goods.
\newblock \emph{European Journal of Operational Research}, 225\penalty0 (1):\penalty0 130--141, 2013.

\bibitem[Gendreau et~al.(1999)Gendreau, Guertin, Potvin, and Taillard]{gendreau1999parallel}
M.~Gendreau, F.~Guertin, J.-Y. Potvin, and E.~Taillard.
\newblock Parallel tabu search for real-time vehicle routing and dispatching.
\newblock \emph{Transportation science}, 33\penalty0 (4):\penalty0 381--390, 1999.

\bibitem[Gendreau et~al.(2006)Gendreau, Guertin, Potvin, and S{\'e}guin]{gendreau2006neighborhood}
M.~Gendreau, F.~Guertin, J.-Y. Potvin, and R.~S{\'e}guin.
\newblock Neighborhood search heuristics for a dynamic vehicle dispatching problem with pick-ups and deliveries.
\newblock \emph{Transportation Research Part C: Emerging Technologies}, 14\penalty0 (3):\penalty0 157--174, 2006.

\bibitem[Ghiani et~al.(2009)Ghiani, Manni, Quaranta, and Triki]{ghiani2009anticipatory}
G.~Ghiani, E.~Manni, A.~Quaranta, and C.~Triki.
\newblock Anticipatory algorithms for same-day courier dispatching.
\newblock \emph{Transportation Research Part E: Logistics and Transportation Review}, 45\penalty0 (1):\penalty0 96--106, 2009.

\bibitem[Ghiani et~al.(2012)Ghiani, Manni, and Thomas]{ghiani2012comparison}
G.~Ghiani, E.~Manni, and B.~W. Thomas.
\newblock A comparison of anticipatory algorithms for the dynamic and stochastic traveling salesman problem.
\newblock \emph{Transportation Science}, 46\penalty0 (3):\penalty0 374--387, 2012.

\bibitem[Heinrich and Silver(2016)]{Heinrich2016}
J.~Heinrich and D.~Silver.
\newblock Deep reinforcement learning from self-play in imperfect-information games.
\newblock \emph{arXiv preprint arXiv:1603.01121}, 2016.

\bibitem[Hvattum et~al.(2006)Hvattum, L{\o}kketangen, and Laporte]{Hvattum06}
L.~Hvattum, A.~L{\o}kketangen, and G.~Laporte.
\newblock Solving a dynamic and stochastic vehicle routing problem with a sample scenario hedging heuristic.
\newblock \emph{Transportation Science}, 40\penalty0 (4):\penalty0 421--438, 2006.

\bibitem[Ichoua et~al.(2000)Ichoua, Gendreau, and Potvin]{ichoua2000diversion}
S.~Ichoua, M.~Gendreau, and J.-Y. Potvin.
\newblock Diversion issues in real-time vehicle dispatching.
\newblock \emph{Transportation Science}, 34\penalty0 (4):\penalty0 426--438, 2000.

\bibitem[Ichoua et~al.(2006)Ichoua, Gendreau, and Potvin]{Ichoua06}
S.~Ichoua, M.~Gendreau, and J.~Potvin.
\newblock Exploiting knowledge about future demands for real-time vehicle dispatching.
\newblock \emph{Transportation Science}, 40\penalty0 (2):\penalty0 211--225, 2006.

\bibitem[Joe and Lau(2020)]{joe20}
W.~Joe and H.~Lau.
\newblock Deep reinforcement learning approach to solve dynamic vehicle routing problem with stochastic customers.
\newblock In J.~Beck, O.~Buffet, J.~Hoffmann, E.~Karpas, and S.~Sohrabj, editors, \emph{Proceedings of the international conference on automated planning and scheduling}, volume~30, pages 394--402, 2020.

\bibitem[Kullman et~al.(2022)Kullman, Cousineau, , Goodson, and Mendoza]{Kullman2019}
N.~Kullman, M.~Cousineau, , J.~Goodson, and J.~Mendoza.
\newblock Dynamic ride-hailing with electric vehicles.
\newblock \emph{Transportation Science}, 56\penalty0 (3):\penalty0 775--794, 2022.

\bibitem[Lample and Chaplot(2017)]{Lample2017}
G.~Lample and D.~Chaplot.
\newblock Playing fps games with deep reinforcement learning.
\newblock In \emph{Proceedings of the Thirty-First AAAI Conference on Artificial Intelligence}, AAAI'17, pages 2140--2146. AAAI Press, 2017.
\newblock URL \url{http://dl.acm.org/citation.cfm?id=3298483.3298548}.

\bibitem[Liu et~al.(2017)Liu, Logan, Liu, Xu, Tang, and Wang]{Liu2017}
Y.~Liu, B.~Logan, N.~Liu, Z.~Xu, J.~Tang, and Y.~Wang.
\newblock Deep reinforcement learning for dynamic treatment regimes on medical registry data.
\newblock In \emph{2017 IEEE International Conference on Healthcare Informatics (ICHI)}, pages 380--385. IEEE, 2017.

\bibitem[Meisel(2011)]{meisel2011anticipatory}
S.~Meisel.
\newblock \emph{Anticipatory Optimization for Dynamic Decision Making}, volume~51 of \emph{Operations Research/Computer Science Interfaces Series}.
\newblock Springer, 2011.

\bibitem[Mitrovi\'{c}-Mini\'{c} and Laporte(2004)]{Mitro04}
S.~Mitrovi\'{c}-Mini\'{c} and G.~Laporte.
\newblock Waiting strategies for the dynamic pickup and delivery problem with time windows.
\newblock \emph{Transportation Research Part B: Methodological}, 38\penalty0 (7):\penalty0 635--655, 2004.

\bibitem[Mnih et~al.(2015)Mnih, Kavukcuoglu, Silver, Rusu, Veness, Bellemare, Graves, Riedmiller, Fidjeland, Ostrovski, et~al.]{Mnih2015}
V.~Mnih, K.~Kavukcuoglu, D.~Silver, A.~Rusu, J.~Veness, M.~Bellemare, A.~Graves, M.~Riedmiller, A.~Fidjeland, G.~Ostrovski, et~al.
\newblock Human-level control through deep reinforcement learning.
\newblock \emph{Nature}, 518\penalty0 (7540):\penalty0 529--533, 2015.

\bibitem[Palombarini and Mart\'{i}nez(2022)]{Palombarini22}
J.~Palombarini and C.~Mart\'{i}nez.
\newblock End-to-end on-line rescheduling from gantt chart images using deep reinforcement learning.
\newblock \emph{International Journal of Production Research}, 60\penalty0 (14):\penalty0 4434--4463, 2022.

\bibitem[Psaraftis(1980)]{psaraftis1980dynamic}
H.~N. Psaraftis.
\newblock A dynamic programming solution to the single vehicle many-to-many immediate request dial-a-ride problem.
\newblock \emph{Transportation Science}, 14\penalty0 (2):\penalty0 130--154, 1980.

\bibitem[{RLlib}(2024)]{rllib}
{RLlib}.
\newblock Industry-grade reinforcement learning, 2024.
\newblock URL \url{https://docs.ray.io/en/latest/rllib/rllib-algorithms.html#dqn}.
\newblock Accessed on June 16, 2024.

\bibitem[Sallab et~al.(2017)Sallab, Abdou, Perot, and Yogamani]{Sallab2017}
A.~Sallab, M.~Abdou, E.~Perot, and S.~Yogamani.
\newblock Deep reinforcement learning framework for autonomous driving.
\newblock \emph{Electronic Imaging}, 2017\penalty0 (19):\penalty0 70--76, 2017.

\bibitem[Silver et~al.(2018)Silver, Hubert, Schrittwieser, Antonoglou, Lai, Guez, Lanctot, Sifre, Kumaran, Graepel, et~al.]{Silver2018}
D.~Silver, T.~Hubert, J.~Schrittwieser, I.~Antonoglou, M.~Lai, A.~Guez, M.~Lanctot, L.~Sifre, D.~Kumaran, T.~Graepel, et~al.
\newblock A general reinforcement learning algorithm that masters chess, shogi, and go through self-play.
\newblock \emph{Science}, 362\penalty0 (6419):\penalty0 1140--1144, 2018.

\bibitem[Soeffker and Ulmer(2019)]{soeffker19}
N.~Soeffker and D.~Ulmer, M.and~Mattfeld.
\newblock Adaptive state space partitioning for dynamic decision processes.
\newblock \emph{Business and Information Systems Engineering}, 61\penalty0 (3):\penalty0 261--275, 2019.

\bibitem[Soeffker et~al.(2022)Soeffker, Ulmer, and Mattfeld]{soeffker2022}
N.~Soeffker, M.~Ulmer, and D.~Mattfeld.
\newblock Stochastic dynamic vehicle routing in the light of prescriptive analytics: A review.
\newblock \emph{European Journal of Operational Research}, 298:\penalty0 801--820, 2022.

\bibitem[Thomas and White~III(2007)]{thomas2007dynamic}
B.~W. Thomas and C.~C. White~III.
\newblock The dynamic shortest path problem with anticipation.
\newblock \emph{European Journal of Operational Research}, 176\penalty0 (2):\penalty0 836--854, 2007.

\bibitem[Ulmer et~al.(2018{\natexlab{a}})Ulmer, Mattfeld, and K{\"o}ster]{ulmer2018a}
M.~Ulmer, D.~Mattfeld, and F.~K{\"o}ster.
\newblock Budgeting time for dynamic vehicle routing with stochastic customer requests.
\newblock \emph{Transportation Science}, 52\penalty0 (1):\penalty0 20--37, 2018{\natexlab{a}}.

\bibitem[Ulmer et~al.(2018{\natexlab{b}})Ulmer, Soeffker, and Mattfeld]{ulmer2018b}
M.~Ulmer, N.~Soeffker, and D.~Mattfeld.
\newblock Value function approximation for dynamic multi-period vehicle routing.
\newblock \emph{European Journal of Operational Research}, 269:\penalty0 883--899, 2018{\natexlab{b}}.

\bibitem[Ulmer et~al.(2018{\natexlab{c}})Ulmer, Goodson, Mattfeld, and Hennig]{ulmer2018offline}
M.~W. Ulmer, J.~C. Goodson, D.~C. Mattfeld, and M.~Hennig.
\newblock Offline--online approximate dynamic programming for dynamic vehicle routing with stochastic requests.
\newblock \emph{Transportation Science}, 53\penalty0 (1):\penalty0 185--202, 2018{\natexlab{c}}.

\bibitem[Ulmer et~al.(2020)Ulmer, Goodson, Mattfeld, and Thomas]{routemdp2020}
M.~W. Ulmer, J.~C. Goodson, D.~C. Mattfeld, and B.~W. Thomas.
\newblock On modeling stochastic dynamic vehicle routing problems.
\newblock \emph{EURO Journal on Transportation and Logistics}, 9\penalty0 (2):\penalty0 100008, 2020.

\bibitem[{van Hemert} and La~Poutr{\'e}(2004)]{van2004dynamic}
J.~I. {van Hemert} and J.~A. La~Poutr{\'e}.
\newblock Dynamic routing problems with fruitful regions: Models and evolutionary computation.
\newblock In \emph{Parallel Problem Solving from Nature-PPSN VIII}, pages 692--701. Springer, 2004.

\bibitem[Vinyals et~al.(2019)Vinyals, Babuschkin, Czarnecki, Mathieu, Dudzik, Chung, Choi, Powell, Ewalds, Georgiev, et~al.]{Vinyals2019}
O.~Vinyals, I.~Babuschkin, W.~Czarnecki, M.~Mathieu, A.~Dudzik, J.~Chung, D.~Choi, R.~Powell, T.~Ewalds, P.~Georgiev, et~al.
\newblock Grandmaster level in starcraft ii using multi-agent reinforcement learning.
\newblock \emph{Nature}, 575\penalty0 (7782):\penalty0 350–354, 2019.

\bibitem[Xinquan and Xuefeng(2023)]{Xinquan23}
W.~Xinquan and Y.~Xuefeng.
\newblock A spatial pyramid pooling-based deep reinforcement learning model for dynamic job-shop scheduling problem.
\newblock \emph{Computers and Operations Research}, 160:\penalty0 106401, 2023.

\end{thebibliography}
		
\end{document}